\def\year{2020}\relax
\documentclass[letterpaper]{article}
\usepackage{aaai20}
\usepackage{times}
\usepackage{helvet}
\usepackage{courier}
\usepackage{graphicx}
\usepackage{color}
\usepackage{multirow}
\usepackage{amsmath}
\usepackage{amssymb}
\usepackage{amsfonts}
\usepackage{amsthm}
\usepackage{algorithm}
\usepackage{algorithmicx}
\usepackage{algpseudocode}

\newcommand{\citet}[1]{\citeauthor{#1} \shortcite{#1}}

\title{Long Short-Term Sample Distillation}

\author{Liang Jiang,\textsuperscript{\rm 1} Zujie Wen,\textsuperscript{\rm 1} Zhongping Liang,\textsuperscript{\rm 1} Yafang Wang,\textsuperscript{\rm 1}\thanks{Corresponding author} Gerard de Melo,\textsuperscript{\rm 2} \\
\bf \Large Zhe Li,\textsuperscript{\rm 1} Liangzhuang Ma,\textsuperscript{\rm 1} Jiaxing Zhang,\textsuperscript{\rm 1} Xiaolong Li,\textsuperscript{\rm 1} Yuan Qi,\textsuperscript{\rm 1} \\
\textsuperscript{\rm 1}AI Department, Ant Financial Services Group, \textsuperscript{\rm 2}Rutgers University \\
\{tianxuan.jl, zujie.wzj, zhongping.lzp, yafang.wyf\}@antfin.com, gdm@demelo.org
}

\begin{document}
\maketitle
\begin{abstract}
In the past decade, there has been substantial progress at training increasingly deep neural networks.
Recent advances within the teacher--student training paradigm have established that information about past training updates show promise as a source of guidance during subsequent training steps.
Based on this notion, in this paper, we propose Long Short-Term Sample Distillation, a novel training policy that simultaneously leverages multiple phases of the previous training process to guide the later training updates to a neural network, while efficiently proceeding in just one single generation pass. With Long Short-Term Sample Distillation, the supervision signal for each sample is decomposed into two parts: a long-term signal and a short-term one. The long-term teacher draws on snapshots from several epochs ago in order to provide steadfast guidance and to guarantee teacher--student differences, while the short-term one yields more up-to-date cues with the goal of enabling higher-quality updates. 
Moreover, the teachers for each sample are unique, such that, overall, the model learns from a very diverse set of teachers. Comprehensive experimental results across a range of vision and NLP tasks demonstrate the effectiveness of this new training method.
\end{abstract}

\section{Introduction}
Our ability to train increasingly deep and increasingly large neural networks has led to substantial progress in AI over the past decade, and a number of techniques have been proposed to address challenges such as  overfitting and the vanishing gradient problem, among others. In recent years, several works have considered the Teacher--Student training paradigm, based on the idea of distilling knowledge from teacher models to guide the optimization of a student model~\cite{bucilua2006model,ba2014deep,hinton2015distilling,czarnecki2017sobolev,zagoruyko2016paying}. The original motivation for this framework was the idea of teaching a small model to mimic the behavior of a larger model so as to speedup the inference and reduce the model size, all while retaining the result quality of the original model. Subsequent work adopted this framework to improve the effectiveness of a student model with identical architecture as the teacher model~\cite{yim2017gift,furlanello2018born}. This is achieved by first training a teacher model and then training a student model with identical architecture but differently initialized parameters, supervised by both the ground truth and the teacher's knowledge. Beyond learning from one single teacher, some studies have shown that learning from multiple teachers yields a better student~\cite{you2017learning,mehak2018knowledge}. Instead of this costly two-stage process, recent work has considered Teacher--Student optimization in a single generation~\cite{laine2016temporal,huang2017snapshot,yang2019snapshot}. The core idea is to consider information about previous training updates to the current model as teacher signals for later training steps of the same neural network in one single generation. 

It has been shown that both teacher--student differences and the quality of the teacher are very important in Teacher--Student optimization \cite{yang2019snapshot}. If student and teacher are very similar, it is impossible for the former to learn from the latter. If the teacher exhibits poor performance, it may introduce noise confusing the student. However, it is difficult to guarantee both teacher--student differences and the quality of the teacher in a single generation. During the course of the training, the predictive quality of the model is expected to become better and better, and thus a high-quality teacher ought to be a fairly recent one, while a dissimilar teacher should rather be far from the student. Previous works rely on just a single teacher, making it hard to simultaneously satisfy these two opposing principles. 

In this paper, we propose a novel training regime named Long Short-Term Sample Distillation (LSTSD), which instead draws on numerous teachers and better leverages knowledge from previous training. In particular, the method decomposes the past history of training updates into long-term knowledge and short-term knowledge to guarantee teacher--student differences while simultaneously ensuring a high quality of the teacher. LSTSD divides the training process into several mini-generations, each of which consists of several training epochs, and each training sample is always guided by two teachers: a long-term teacher and a short-term one. The long-term teacher signal comes from the last mini-generation and remains fixed during the course of a mini-generation, so as to provide a steadfast teacher signal and guarantee teacher--student differences. The short-term teacher, in contrast, comes from the previous epoch and changes at every epoch, so as to provide more up-to-date signals that are likely to be of higher quality.
Additionally, motivated by~\citet{you2017learning}, we conjecture that learning from numerous past snapshots from the previous training process leads to a better model. In our method, teacher signals for each sample come from different snapshots in the previous training process, and thus the model learns from a very diverse set teachers at the same time. 

Specifically, in each epoch, we save the probability distribution produced by the corresponding snapshot for each sample when it is selected as training data to update the neural network. This will serve as the short-term teacher in the next epoch, and remain up-to-date at every epoch. Besides the short-term teacher, in the last epoch of a mini-generation, we further save the probability distribution produced by the corresponding snapshot for each sample when it is selected to update the neural network. This will serve as the long-term teacher for the same sample when it is selected to update the model in the next mini-generation, and remains fixed within that mini-generation. 

We conducted experiments across a range of different vision and NLP tasks with a diverse set of neural network architectures to verify the effectiveness and generalization ability of LSTSD. The experimental results demonstrate that LSTSD can improve the performance significantly and can generalize to many different tasks.

\section{Related Work}
In recent years, important advances in artificial intelligence have arisen simply from our ability to train models with more layers and parameters. To address the computational overhead of larger models, techniques such as deep compression~\cite{han2015deep} have been proposed. 
To address the optimization challenges of training increasingly deep neural networks, a number of techniques have been proposed as well. For instance, residual networks~\cite{he2016deep} were proposed to alleviate the problem of vanishing gradients, and dropout~\cite{srivastava2014dropout} was proposed to reduce overfitting. 

In recent years, the Teacher--Student framework has shown great potential for accelerating the inference and improving the performance of neural networks. In this framework, the target model is supervised not only by the ground truth, but also by signals from a teacher model, which aims to help optimize the target model. The Teacher--Student framework was originally proposed to distill knowledge from a large teacher model and guide the training of a small student model, such that the small student model can approximate the result quality of the large model while allowing for inference on resource-constrained devices such as cellphones. In their pioneering work, \citet{bucilua2006model} proposed to distill an ensemble of neural networks into a small neural network to accelerate the model. In many following works, the student model was taught to mimic the behavior of the teacher model by approximating the output or the internal state of the pre-trained teacher model. For instance, in \citet{hinton2015distilling}, the student model was trained to not only predict the ground truth label accurately, but also to produce a softmax distribution matching that of the teacher model as closely as possible. Instead of mimicking the output of the teacher model, \citet{romero2014fitnets} proposed a method in which the student mimics the hidden layers of the teacher model.

Besides distilling a large teacher into a small student for accelerated inference on the network, subsequent studies have found that distilling a teacher into a student model of identical architecture also shows promise. In~\citet{yim2017gift}, a student model achieved faster convergence and greater accuracy by matching the hidden layers with those of a teacher model with identical architecture. \citet{furlanello2018born} proposed born-again networks, in which a re-initialized student learns from a pre-trained teacher of identical architecture, achieving better performance. Beyond learning from single teacher, \citet{you2017learning} showed that learning from multiple teachers leads to a better student. In their work, multiple teachers are combined via a voting strategy, and the student is required to mimic both the internal layers and outputs of multiple teachers.

All of the aforementioned Teacher--Student methods divided the overall training process into multiple generations: the teacher and the student generations. In the teacher generation, a teacher model is pre-trained, while in the student generation, a student model is trained, supervised by the pre-trained teacher model.  This training regime however entails an additional computational burden, because a series of models need to be optimized one by one. To reduce the extra computational overhead, several methods have been proposed to implement Teacher--Student Optimization in one single generation. In these methods, information distilled from the previous training process serves as a teacher signal for subsequent training of the same generation. \citet{tarvainen2017mean} proposed the Mean Teacher approach, in which the moving average parameters of all snapshots in the previous training process is used as a teacher for later training of the same generation. \citet{yang2019snapshot} proposed Snapshot Distillation, in which a training generation is divided into several mini-generations. During the training of each mini-generation, the parameters of the last snapshot model in the previous mini-generation serve as a teacher model. In Temporal Ensembles, for each sample, the teacher signal is the moving average probability produced by the snapshots when the sample was selected as training data in all previous epochs~\cite{laine2016temporal}. 

In this work, we propose Long Short-Term Sample Distillation to obtain better sample-level Teacher--Student optimization in one generation. With Long Short-Term Sample Distillation, the teacher signal comes from two teachers: a long-term teacher and a short-term one. The long-term teacher comes from the previous mini-generation and remains fixed within the range of the next mini-generation, aiming to provide a stable teacher signal and guarantee teacher--student differences. The short-term teacher comes from the previous epoch and remains fixed only within the next epoch, aiming to provide a more up-to-date teacher signal guaranteeing the teacher quality. It is worth mentioning that the teacher signals of each sample are produced by the snapshot when the sample was selected as training data. Thus, each sample has unique teachers, enabling the model to learn from numerous teachers at the same time.

\begin{figure*}[t]
\centering
\includegraphics[width=0.95\textwidth]{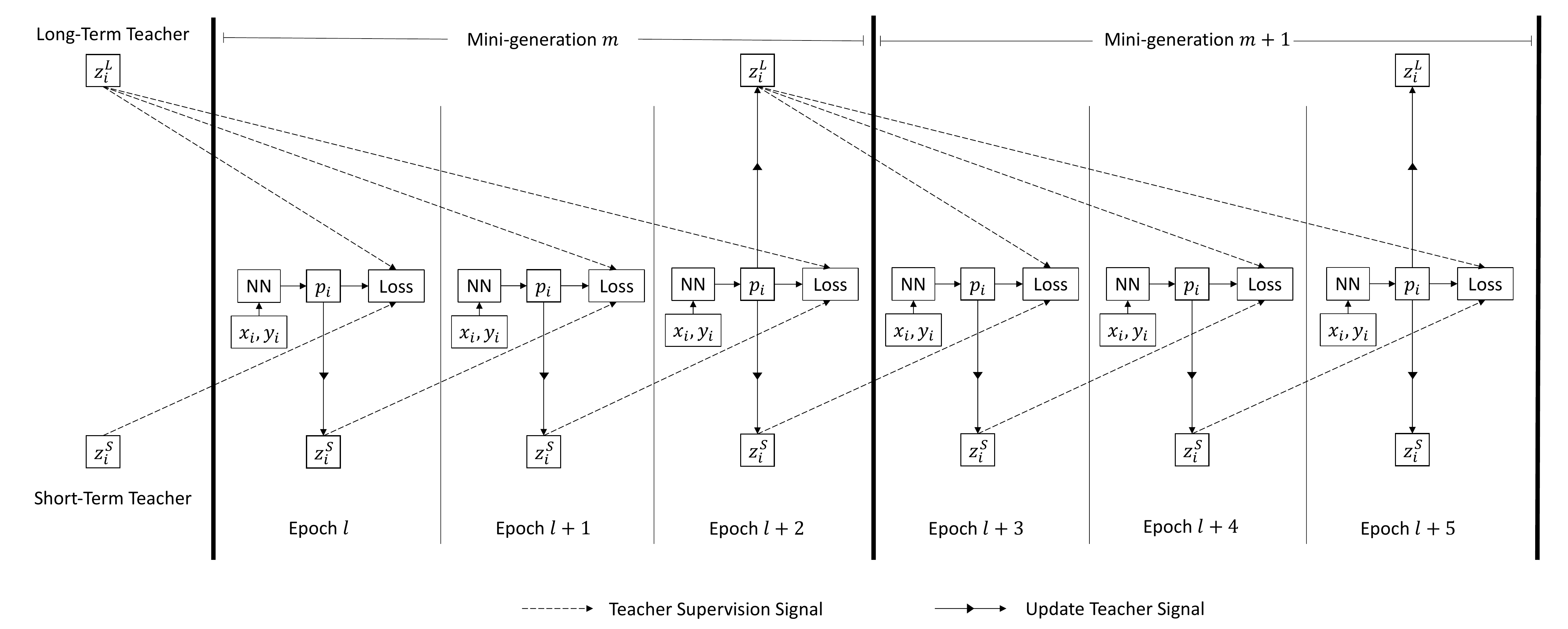}
\caption{An illustration of Long Short-Term Sample Distillation. Here, we assume each mini-generation includes 3 training epochs. \textit{NN} denotes the neural network to optimize, and \textit{Loss} denotes the loss function including cross-entropy, long-term teacher loss, and short-term teacher loss.}
\label{sd_framework}
\end{figure*}

\section{Method}
In this section, we introduce our proposed Long Short-Term Sample Distillation (LSTSD) in detail. 
For background, we shall first briefly review Mini-Batch SGD Optimization, Teacher--Student Optimization, and One-Generation Teacher-Student Optimization. Subsequently, we will describe our novel Long Short-Term Sample Distillation approach.

\subsection{Mini-Batch SGD Optimization}
Consider a classification problem optimized with mini-batch SGD. We have a training dataset consisting of samples and labels $(x,y)\in\mathcal{X}\times\mathcal{Y}$. Our goal is to find a function $f(x; \theta):\mathcal{X}\rightarrow\mathcal{Y}$ that generalizes well to unseen data, where $f(x)$ is often a deep neural network parameterized by $\theta$. One of the most widely used methods to learn $\theta$ is by minimizing the cross-entropy between the predicted probability distribution and the ground truth using mini-batch SGD. 

Specifically, given a dataset with $N$ samples $\mathcal{D}=\{(x_1,y_1),(x_2,y_2),...,(x_N,y_N) \mid (x_i,y_i)\in\mathcal{X}\times{\mathcal{Y}}\}$, we define the objective to optimize for as the cross-entropy, i.e., 
$$
\mathcal{L}=-\frac{1}{|\mathcal{D}|}\sum_{(x_i,y_i)\in \mathcal{D}} \ln f_{y_i}(x_i; \theta),
$$
\noindent where $f_{y_i}$ denotes the probability of the label $y_i$ predicted by the neural network.

To find good local optima of $f(x; \theta)$ that generalize well to unseen data, mini-batch SGD is usually invoked to minimize the objective $\mathcal{L}$. Specifically, in the $t$-th iteration, a mini-batch $\mathcal{B}$ is randomly sampled to train the model $f(x;\theta_{t})$. First, we determine the objective of $f(x;\theta_{t})$ on $\mathcal{B}$,
$$
\mathcal{L}({\mathcal{B}};\theta_{t})=-\frac{1}{|\mathcal{B}|}\sum_{(x_i,y_i)\in\mathcal{B}} \ln f_{y_i}(x_i; \theta_{t}).
$$
\noindent Then, we compute the gradient of $\mathcal{L}(\mathcal{B};\theta_t)$ with respect to $\theta$, and adjust each parameter in $\theta$ in the direction of gradients, 
$$
\theta_{t+1}=\theta_{t}-\eta\frac{\partial \mathcal{L}(\mathcal{B};\theta_{t})}{\partial\theta},
$$
\noindent where $\eta$ denotes the learning rate. At this point, the $t$-th iteration of optimization is completed. We simply repeat this procedure until some predefined stopping criteria are fulfilled, in order to obtain the sought optimal parameters $\theta^*$.

\subsection{Teacher--Student Optimization}
The process of SGD searches over the parameter space to find a $\theta^*$ that best fits the given dataset $\mathcal{D}$. However, as the depth of neural networks and the amount of parameters increases, $\theta^*$ often overfits $\mathcal{D}$. \citet{guo2017calibration} found that this may stem from the fact that the supervision is provided as one-hot vectors, which forces the network to overwhelmingly prefer the true class over all other classes. This is often not an optimal choice because rich information of class-level similarity is simply discarded. 

One of the methods to address this issue is the Teacher--Student framework, where a teacher model provides complementary information to help the training of student model. Specifically, the objective of the student model is now not only to predict the ground truth label of each sample correctly, but also to mimic the behavior of the teacher model. One way to mimic the teacher is to approximate the probability distribution produced by the teacher. This is usually achieved by adding an extra term to minimize the divergence between the probability distributions of the teacher model and student model. The loss function of the student model can be formulated as
\begin{align*}
    \mathcal{L}(\mathcal{B};\theta_t)=&-\frac{1}{|\mathcal{B}|}\sum_{(x_i,y_i)\in \mathcal{B}}\big[\,\ln f_{y_i}(x_i; \theta_t)\\
    &+\lambda\, \mathrm{KL}[f(x_i; \theta)\,||\,f^T(x_i;\theta^T)]\,\,\big],
\end{align*}
\noindent where $f^T$ denotes the teacher network parameterized by $\theta^T$, and $\mathrm{KL}$ denotes the KL-divergence function measuring the divergence between the probability distributions of the teacher and  student models. In the Teacher--Student framework, besides the one-hot vector of the ground truth label, the student model receives the probability distribution of the teacher model as an additional form of supervision that is much smoother than a one-hot vector and may mitigate the problem of overfitting. 

In the flow chart of the Teacher--Student framework, the overall training process is usually divided into two generations: the teacher generation and the student generation, to train the teacher model and student model, respectively. However, this brings an additional computational time cost to the training process. To alleviate this issue, several methods have been proposed to implement Teacher--Student Optimization in one generation, which we shall refer to as One-Generation Teacher-Student Optimization.

\subsection{One-Generation Teacher--Student Optimization}
In One-Generation Teacher--Student Optimization, there is no need to pre-train a distinct teacher model, as the teacher signals comes from the previous training process of the same generation instead. Specifically, suppose that at the $t$-th step, a mini-batch $\mathcal{B}$ is sampled to train the model. For each sample $x_i$ in the mini-batch data $\mathcal{B}$, the supervision signal contains the ground truth label $y_i$ and the probability distribution of $x_i$ produced by the teacher snapshot $S_i$.
\begin{align*}
\mathcal{L}(\mathcal{B};\theta_{t})=&-\frac{1}{|\mathcal{B}|}\sum_{(x_i,y_i)\in\mathcal{B}}\big[\,\ln f_{y_i}(x_i;\theta_{t}) \\
&+\lambda\, \mathrm{KL}[f(x_i;\theta_{t})||f(x_i;\tilde{\theta}_i)]\,\,\big],
\end{align*}
\noindent where $\theta_{t}$ denotes parameters in the neural network at the $t$-th time step, $\tilde{\theta}_i$ denotes the parameters in the teacher snapshot $S_i$ of sample $x_i$, which is a snapshot model at some time step in the previous training process. 

The key question for One-Generation Teacher--Student Optimization is how to choose the teacher snapshot for each sample: Should we use one teacher for all samples or unique teachers for each sample? Should we use a snapshot far in the past or near the present? In this work, we investigate these two problems and propose Long Short-Term Sample Distillation to obtain better One-Generation Teacher-Student Optimization.

\subsection{Long Short-Term Sample Distillation}
In our proposed LSTSD method, each sample has two unique teachers: a long-term teacher and a short-term one. The long-term teacher for a sample is the snapshot model when it was selected as training data in the last epoch of the previous mini-generation, and remains fixed in the next mini-generation. The short-term teacher for a sample is the snapshot model when it was selected as training data in the previous epoch, and is updated at every epoch. 

\paragraph{Short-Term Teacher.}
As illustrated in Figure~\ref{sd_framework}, in the $l$-th epoch of the $m$-th mini-generation, the dataset $\mathcal{D}$ is shuffled to ensure that samples are ordered randomly, which is denoted by $\mathcal{D}^l$, and the model is trained with mini-batches sampled from $\mathcal{D}^l$ sequentially. Suppose at the $r$-th step, data $(x_i,y_i)$ is selected as training data to update parameters $\theta_{r}^l$ in the corresponding snapshot model $S^l_{r}$. Then, $S^l_{r}$ is used as the short-term teacher for $(x_i,y_i)$ in the $(l+1)$-th epoch. Instead of saving $\theta_{r}^l$, we maintain a short-term teacher vector $z^S$ to retain the probability distribution of $(x_i,y_i)$,
$$
z_i^S=p(x_i)=f(x_i;\theta_{r}^l),
$$
\noindent where we use $z_i^S$ to denote the short-term teacher vector $z^S[x_i]$ of $x_i$ for clarity. Storing the probability distribution instead of the parameters eliminates the extra computational cost entailed by calculating the probability repeatedly in the $(l+1)$-th epoch. After the $l$-th epoch of training has completed, the short-term teacher vector $z^S$, which contains knowledge of all snapshots in the $l$-th epoch, will be used as the short-term teacher in the $(l+1)$-th epoch. The short-term teacher is updated at every epoch to remain up-to-date.

\paragraph{Long-Term Teacher.}
At the beginning of the last epoch in the $m$-th mini-generation (i.e., the $(l+2)$-th epoch in Figure~\ref{sd_framework}), the training dataset $\mathcal{D}$ is shuffled again into $\mathcal{D}^{l+2}$. Suppose that at the $w$-th step, data $(x_i,y_i)$ is selected as training data to update the parameters $\theta_{w}^{l+2}$ in the corresponding snapshot model $S^{l+2}_{w}$. Then, $S^{l+2}_{w}$ will serve as the long-term teacher for $(x_i,y_i)$ and remain fixed in the $(m+1)$-th mini-generation. Instead of storing $\theta_{w}^{l+2}$, we maintain a long-term teacher vector $z^L$ to capture the probability distribution of $(x_i,y_i)$,
$$
z_i^L=p(x_i)=f(x_i;\theta_{w}^{l+2}),
$$
\noindent where we use $z_i^L$ to denote the long-term teacher vector $z^L[x_i]$ of $x_i$ for clarity. After the $m$-th mini-generation of training has completed, the long-term teacher vector $z^L$, which contains knowledge of all snapshots in the last epoch of the $m$-th mini-generation, will be used as long-term teachers in the $(m+1)$-th mini-generation. The long-term teacher is updated only in the last epoch of every mini-generation, and remains unchanged in other epochs.
\begin{algorithm}
\caption{Long Short-Term Sample Distillation}
%\small{}
\begin{algorithmic}[1]
    \Require $\mathcal{D}$ = Training set
    \Require $M$ = Number of mini-generations
    \Require $E$ = Number of epochs in each mini-generation
    \Require $\lambda^L$ = Weight of long-term distillation loss
    \Require $\lambda^S$ = Weight of short-term distillation loss
    \Require $f(x;\theta)$ = neural network parameterized by $\theta$
    \For{$m=1$ to $M$}
        \For{$e=1$ to $E$}
            \State $\tilde{\mathcal{D}} \gets$ shuffle training set  $\mathcal{D}$
            \For{each mini-batch $\mathcal{B}$ in $\tilde{\mathcal{D}}$}
                \State $\mathcal{L}^C \gets -\frac{1}{|\mathcal{B}|}\sum_{(x_i,y_i)\in\mathcal{B}}\ln f_{y_i}(x_i;\theta)$
                \If{$m>1$}
                    \State $\mathcal{L}^L\gets \frac{1}{|\mathcal{B}|}\sum_{(x_i,y_i)\in\mathcal{B}}\mathrm{KL}[f(x_i;\theta)||z^L_i]$ 
                    \State $\mathcal{L}^S\gets \frac{1}{|\mathcal{B}|}\sum_{(x_i,y_i)\in\mathcal{B}}\mathrm{KL}[f(x_i;\theta)||z^S_i]$
                    \State $\mathcal{L} \gets \mathcal{L}^C+\lambda^L\mathcal{L}^L+\lambda^S\mathcal{L}^S$
                \Else
                    \State $\mathcal{L}\gets\mathcal{L}^C$
                \EndIf
                \State Update $\theta$ using gradient of $\mathcal{L}$
                \State $z_{i\in \mathcal{B}}^S \gets f(\{x_i\in\mathcal{B}\};\theta)$
                \If{e=E}
                    \State $z_{i\in \mathcal{B}}^L \gets f(\{x_i\in \mathcal{B}\};\theta)$
                \EndIf
            \EndFor
        \EndFor
    \EndFor
\end{algorithmic}
\label{lstsd_algo}
\end{algorithm}
\paragraph{Long Short-Term Teacher-Student Optimization.} In the $(m+1)$-th mini-generation, besides the ground truth supervision, each sample is provided a long-term teacher $z^L$ from the previous mini-generation and a short-term teacher $z^S$ from the previous epoch as described above. Therefore, the model is required to not only correctly predict the ground truth label, but also to simultaneously approximate the probability distributions of the long-term teacher and short-term teacher. Without loss of generality, let us consider the second epoch in the $(m+1)$-th epoch, i.e., the $(l+4)$-th epoch in Figure~\ref{sd_framework}. The short-term teacher $z^S$ comes from the $(l+3)$-th epoch, and the long-term teacher $z^L$ comes from the $(l+2)$-th epoch. At the beginning of the $(l+4)$-th epoch, the dataset $\mathcal{D}$ is shuffled into $\mathcal{D}^{l+4}$. Suppose at the $t$-th step, a mini-batch $\mathcal{B}$ was sampled from $\mathcal{D}^{l+4}$ to update the parameters. The supervision signals of each sample $x_i\in \mathcal{B}$ consists of three components: the ground truth label $y_i$, the long-term teacher signal $z^L_i$, and the short-term teacher signal $z^S_i$. The training objective of $\mathcal{B}$ can be formulated as
%\begin{small}
\begin{equation}
\begin{aligned}
\mathcal{L}&=\mathcal{L}^C+\lambda^L\,\mathcal{L}^L+\lambda^S\,\mathcal{L}^S \\
\mathcal{L}^C&=-\frac{1}{|\mathcal{B}|}\sum_{(x_i,y_i)\in \mathcal{B}}\ln f_{y_i}(x_i;\theta^{l+4}_{t}) \\
\mathcal{L}^L&=\frac{1}{|\mathcal{B}|}\sum_{(x_i,y_i)\in\mathcal{B}}\mathrm{KL}[f(x_i;\theta^{l+4}_{t})||z^L_i] \\
\mathcal{L}^S&=\frac{1}{|\mathcal{B}|}\sum_{(x_i,y_i)\in\mathcal{B}}\mathrm{KL}[f(x_i;\theta^{l+4}_{t})||z^S_i],
\end{aligned}
\label{sd_eq}
\end{equation}
%\end{small}
Here, $\lambda^L$ and $\lambda^S$ denote the weight of the long-term teacher signal and short-term teacher signal, respectively, $\theta^{l+4}_{t}$ denotes the parameters in the corresponding snapshot model at the $t$-th step in the $(l+4)$-th epoch, and $z_{i}^L$, $z_{i}^S$ represent the long-term teacher signal and short-term teacher signal for sample $x_i$, respectively. The LSTSD procedure is given more formally as Algorithm~\ref{lstsd_algo}.

As indicated by Equation~\ref{sd_eq}, each sample in $\mathcal{D}^{l+4}_t$ has two teacher snapshots from the previous training process. The long-term teacher provides a stable signal establishing teacher--student differences, and the short-term teacher provides a more up-to-date signal guaranteeing the quality of teacher. Since the teachers for each sample are unique, and  $\mathcal{D}^{l+4}$, $\mathcal{D}^{l+3}$, $\mathcal{D}^{l+2}$ contain the same samples but in different order, the $t$-th mini-batch in $\mathcal{D}^{l+4}$ contains samples from different batches in $\mathcal{D}^{l+3}$ and $\mathcal{D}^{l+2}$. Thus, the model may learn from numerous long-term teachers and short-term teachers at the same time. Furthermore, since we always save the probability distribution of the samples rather than the parameters in teacher snapshots, there is no need to calculate the probability of teacher snapshots repeatedly. LSTSD brings almost no extra computational cost, making it more widely applicable in a variety of settings. 

\section{Experiments}

\begin{table*}[t]
\centering
\caption{CIFAR100 classification accuracy ($\%$) obtained by different networks. Bold values indicate the best performance.}
\begin{tabular}{|c|c|c|c|c|c|} \hline
\textbf{Network} & ResNet-20 & ResNet-32 & ResNet-56 & ResNet-110 & DenseNet-100\\ \hline\hline
Vanilla & 66.43 & 68.39 & 70.06 & 71.47 & 78.00\\
\hline
Mean Teacher & 68.37 & 70.26 & 72.00 & 72.57 & 76.80\\
Snapshot Ensembles & 67.46 & 69.49 & 70.45 & 71.91 & 78.00\\
Temporal Ensembles & 67.90 &  69.79 & 71.20 & 71.99 & 77.13\\
Snapshot Distillation & 68.24 & 69.84 & 70.78 & 72.48 & 78.83 \\
\hline
LSTSD & \textbf{69.42} & \textbf{71.51} & \textbf{73.17} & \textbf{73.83} & \textbf{79.35}\\
\hline
\end{tabular}
\label{cifar100_result}
\end{table*}

To verify the effectiveness and generalization ability of our proposed Long Short-Term Sample Distillation technique, we conducted a comprehensive series of experiments with different neural network architectures on both vision and NLP tasks. In this section, we introduce the baselines, experimental settings, and analyze the experimental results.
\begin{table*}[t]
\centering
\caption{GLUE results ($\%$) obtained by BERT and CNN, the metric of RTE MPRC, and SST-2 is accuracy, and the metrics of CoLA is Matthew's Corr. Bold values indicate the best performance.}
\begin{tabular}{|c|c|c|c|c|c|} \hline
\textbf{Network} & \textbf{Method} & \textbf{RTE} & \textbf{MRPC} & \textbf{SST-2} & \textbf{CoLA} 
\\ \hline\hline
\multirow{5}{*}{BERT} & Vanilla & 72.20 & 86.03 & 93.00 & 58.54\\
& Mean Teacher & 70.39 & 85.29 & 92.89 & \textbf{61.75}\\
& Snapshot Ensembles & 73.29 & 86.76 & 92.32 & 59.53\\
& Temporal Ensembles & 71.50 & 85.78 & 93.11 & 60.56\\
& Snapshot Distillation & 74.01 & 87.25 & 93.12 & 60.09\\
& LSTSD & \textbf{74.73} & \textbf{89.22} & \textbf{93.35} & 61.59\\
\hline
\multirow{6}{*}{CNN} & Vanilla & 53.79 & 70.83 & 70.99 & 9.70\\
& Mean Teacher & 54.87 & 71.81 & 70.41 & 9.32\\
& Snapshot Ensembles & 55.60 & 70.83 & 71.67 & 10.51\\
& Temporal Ensembles & 54.87 & 72.06 & 70.53 & 11.27 \\
& Snapshot Distillation & 56.68 & \textbf{73.77} & 71.67 & 12.81\\
& LSTSD & \textbf{57.40} & 73.28 & \textbf{72.36} & \textbf{14.50}\\
\hline
\end{tabular}
\label{glue_result}
\end{table*}

\subsection{Baselines}
To evaluate our proposed LSTSD, we compared it with Mean Teacher~\cite{tarvainen2017mean}, Temporal Ensembles~\cite{laine2016temporal}, Snapshot Ensembles~\cite{huang2017snapshot} and Snapshot Distillation~\cite{yang2019snapshot}.
%which are all methods leveraging knowledge from the previous training process in one generation. 

The \textbf{Mean Teacher} approach generates the teacher model by calculating the moving weighted average parameters over all training steps, aiming to produce a more accurate teacher model than using the final weights directly and allowing the model to learn from all snapshots in previous training steps. Specifically, the parameters in the teacher model are computed as $\theta_{t+1}^{'}=\alpha\theta_t^{'}+(1-\alpha)\theta_t$ at the $t$-th iteration. As suggested by the original paper, we set $\alpha=0.999$.

\textbf{Temporal Ensembles} saves each sample's moving average probability produced by the neural network when the sample was selected as training data to update the parameters in the previous training process, rather than saving the parameters of the neural network. Specifically, the moving average probability is computed as $Z=\alpha Z+(1-\alpha)z$ at every epoch, where $Z$ denotes the moving average probability and $z$ denotes the probability at the current time step. As suggested by the original paper, we set $\alpha=0.6$.

\textbf{Snapshot Ensembles} divides the  training process into several mini-generations, in each of which the model is trained with a cyclic learning rate to force the model to converge to different well-performing local minima. After training, the last snapshots in each mini-generation are ensembled to boost the performance.

Similar to Snapshot Ensembles, \textbf{Snapshot Distillation} also divides the overall training process into several mini-generations. In each mini-generation, the last snapshot in the previous mini-generation is used as a teacher. To assure a difference between student and teacher, a cyclic learning rate is applied in each mini-generation.

\subsection{Experimental Setup}

We applied all methods to ResNets and DenseNets for vision tasks, and to CNNs and BERT for NLP tasks. 
%and compare their results to evaluate LSTSD. 
For all baselines, we used the hyperparameters mentioned above, and for LSTSD, we set each mini-generation to $6$ epochs. 
To better understand LSTSD, besides comparing it with these baselines, we also conducted experiments on several variants of LSTSD to measure the influence of long-term teacher, short-term teacher and numerous teachers separately. Also, we did a sensitivity analysis on the length of mini-generation, to investigate the influence on performance of different length of mini-generation.

\begin{table*}[t]
\centering
\caption{CIFAR100 classification accuracy ($\%$) obtained by different variants of Long Short-Term Sample Distillation. Values in parentheses after each result represent the absolute difference to LSTSD.}
\begin{tabular}{|c|c|c|c|c|} \hline
\textbf{Network} & ResNet-20 & ResNet-32 & ResNet-56 & ResNet-110 \\ \hline\hline
LSTSD & 69.42 (-0.00) & 71.51 (-0.00) & 73.17 (-0.00) & 73.83 (-0.00)\\
LSTSD (w/o Long) & 69.09 (-0.34) & 71.16 (-0.35) & 73.15 (-0.02) & 73.35 (-0.48) \\
LSTSD (w/o Short) & 68.82 (-0.60) & 70.71 (-0.80) & 72.79 (-0.38) & 73.23 (-0.60) \\
LSTSD (single) & 67.85 (-1.57) & 69.88 (-1.63) & 70.66 (-2.51) & 72.25 (-1.58)\\
\hline
\end{tabular}
\label{ablation_result}
\end{table*}

\paragraph{Computer Vision.}
For vision, we evaluate LSTSD on the CIFAR100 dataset, which contains 60,000 RGB images of $32\times32$ size, split into a training set of 50,000 images and a testing set of 10,000 images. The images are uniformly distributed over all 100 labels, examples of which include \emph{bottle}, \emph{bed}, \emph{clock}, and \emph{apple}. We investigate two groups of baseline models. The first group contains ResNets with different numbers of layers (20, 32, 56, 110) as baseline backbones, with architectures matching those of \citet{he2016deep}. The second group contains DenseNets with 100 layers, in which the base feature length and growth rate are 24 and 80, respectively~\cite{huang2017densely}. ResNets are trained for 164 epochs with a batch size of $128$, while DenseNets are trained for 300 epochs with a batch size of $64$. We trained both ResNets and DenseNets using SGD with a weight decay of $0.0001$, a Nesterov momentum of 0.9 and a base learning rate of $0.1$, which was divided by $10$ at the $25\%$, $50\%$, $75\%$ of the training process.

Standard data augmentation was applied in the training process, i.e., each image was symmetrically-padded with a 4-pixel margin on each of the four sides. In the enlarged $40\times40$ image, a subregion with $32\times32$ pixels is randomly cropped and flipped with a probability of $0.5$. 

We set the length of each mini-generation to 40 epochs for Snapshot Ensembles and Snapshot Distillation following~\citet{yang2019snapshot}. The best weights of the teacher loss for all baselines were determined by grid search. We found the best $\lambda^S=4.0$, $\lambda^L=2.4$ and length of mini-generation to 6 epochs for LSTSD using a residual network of 20 layers with grid search, and used the same setting for other network backbones. Following~\citet{hinton2015distilling}, we divided the teacher  and student signal (in \textit{logits}, the neural responses before the soft-max) by a temperature coefficient $T=2$ in calculating the distilling losses, which has been proven effective to soften the teacher signal and student signal in Teacher--Student Optimization.

\paragraph{Natural Language Processing.}
For NLP, we used the well-known GLUE benchmark data~\cite{wang2019glue}, which is a collection of diverse natural language understanding tasks,  including question answering, sentiment analysis, text similarity, and textual entailment. Among all datasets in GLUE, we selected several classification datasets to conduct experiments, including RTE, MRPC, CoLA, and SST-2. 
We used BERT~\cite{devlin2018bert} and CNNs as baseline backbones. BERT has 12 layers, each of which has 12 self-attention heads with the hidden layer size set to 768. We initialized BERT with the parameters provided in \citet{devlin2018bert}, which were trained with a Masked Language Model (MLM) objective on a large unannotated corpus. We optimized BERT using Adam for 50 epochs, with the base learning rate set to $5e-5$ and batch size set to 64. We initialized CNNs randomly and optimized them using SGD for 50 epochs with a learning rate of $8e-3$ and a batch size of 32. We set the temperature to $1$, since there are only a few classes in datasets of GLUE, the probability distributions are much smoother than in datasets with a large number of classes, and no further softening is needed.

\subsection{Experimental Results}
\paragraph{Computer Vision.} On vision tasks, as shown in Table~\ref{cifar100_result}, LSTSD brings consistent accuracy gains for all models, regardless of network backbones. Specifically, LSTSD achieves accuracies of $69.42\%$, $71.51\%$, $73.17\%$, $73.83\%$ for residual networks with 20, 32, 56 and 110 layers, respectively, and  $79.35\%$ for DenseNet-100.

All methods outperform the vanilla networks of all layers, which demonstrates the effectiveness of introducing either long-term knowledge or short-term knowledge from the previous training process of the same generation to help the optimization of neural networks. In Temporal Ensembles, the teacher signal quickly decays by 0.6 per epoch, which made it more like a short-term signal guaranteeing the quality of teacher. In the Mean Teacher approach, the teacher signal decays by 0.999 at every iteration, which amounts to about 0.6 per epoch on a dataset with 500 iterations. Thus, Mean Teacher is also more like a short-term teacher. Moreover, the teacher signal in Snapshot Distillation remains fixed in each mini-generation, which made it more like a long-term signal guaranteeing teacher--student differences. The fact that LSTSD outperforms all of these demonstrates the advantage of decomposing the teacher signal into a long-term and short-term signals and leveraging both simultaneously.

\paragraph{Natural Language Processing.} On NLP tasks, as shown in Table~\ref{glue_result}, LSTSD also outperforms other methods on the four datasets. Specifically, when applied to BERT, LSTSD achieves accuracies of $74.73\%$, $89.22\%$, $93.35\%$ on RTE, MRPC, and SST-2, respectively, outperforming all other baselines and vanilla BERT. It achieves a Matthew's Correlation of $61.59\%$ on CoLA, which is comparable with Mean Teacher. Similarly, when applied to CNN, LSTSD outperforms all baselines and vanilla CNNs on RTE, SST-2, CoLA, and is comparable to Snapshot Distillation on MRPC. It is worth mentioning that BERT and CNN are substantially different architectures, since the core of BERT is an attention mechanism, while the core of CNN are convolutions. Despite the great difference between BERT and CNN, LSTSD achieves consistent gains with both of them, which further establishes the generalization ability of LSTSD.

\paragraph{Analysis of Model Variants.}
To better understand Long Short-Term Sample Distillation, we conducted experiments on  CIFAR100 using ResNet-20 to evaluate the importance of the long-term teacher and short-term teacher on the performance separately. Specifically, we set $\lambda^L=0$ and $\lambda^S=4.0$, in order to evaluate the importance of the long-term teacher signal, denoted by \emph{LSTSD (w/o Long)}. Similarly, we evaluate the importance of the short-term teacher signal by setting $\lambda^L=2.4$ and $\lambda^S=0$, denoted by \emph{LSTSD (w/o Short)}. As shown in Table~\ref{ablation_result}, eliminating long-term knowledge or short-term knowledge degrades the performance significantly, suggesting that it is necessary to leverage both long-term and short-term knowledge jointly.

In Long Short-Term Sample Distillation, each sample has unique teachers, enabling the model to learn from numerous teachers. To validate whether the model benefits from numerous teachers, we compare LSTSD with a variant in which all samples learn from a single teacher. Specifically, rather than taking the snapshot when a sample was selected as training data as the teacher model for the sample, we use the last snapshot in the previous mini-generation as the long-term teacher and the last snapshot in the previous epoch as the short-term teacher, such that all samples share the same long-term teacher and short-term teacher in every epoch (denoting this method as \emph{LSTSD (single)} in~Table \ref{ablation_result}). The comparison between LSTSD (single) and LSTSD shows that replacing numerous teachers with one teacher degrades the performance significantly, which shows the advantage of learning from numerous teachers at the same time.

\begin{figure}[t]
\centering
\includegraphics[width=0.95\columnwidth]{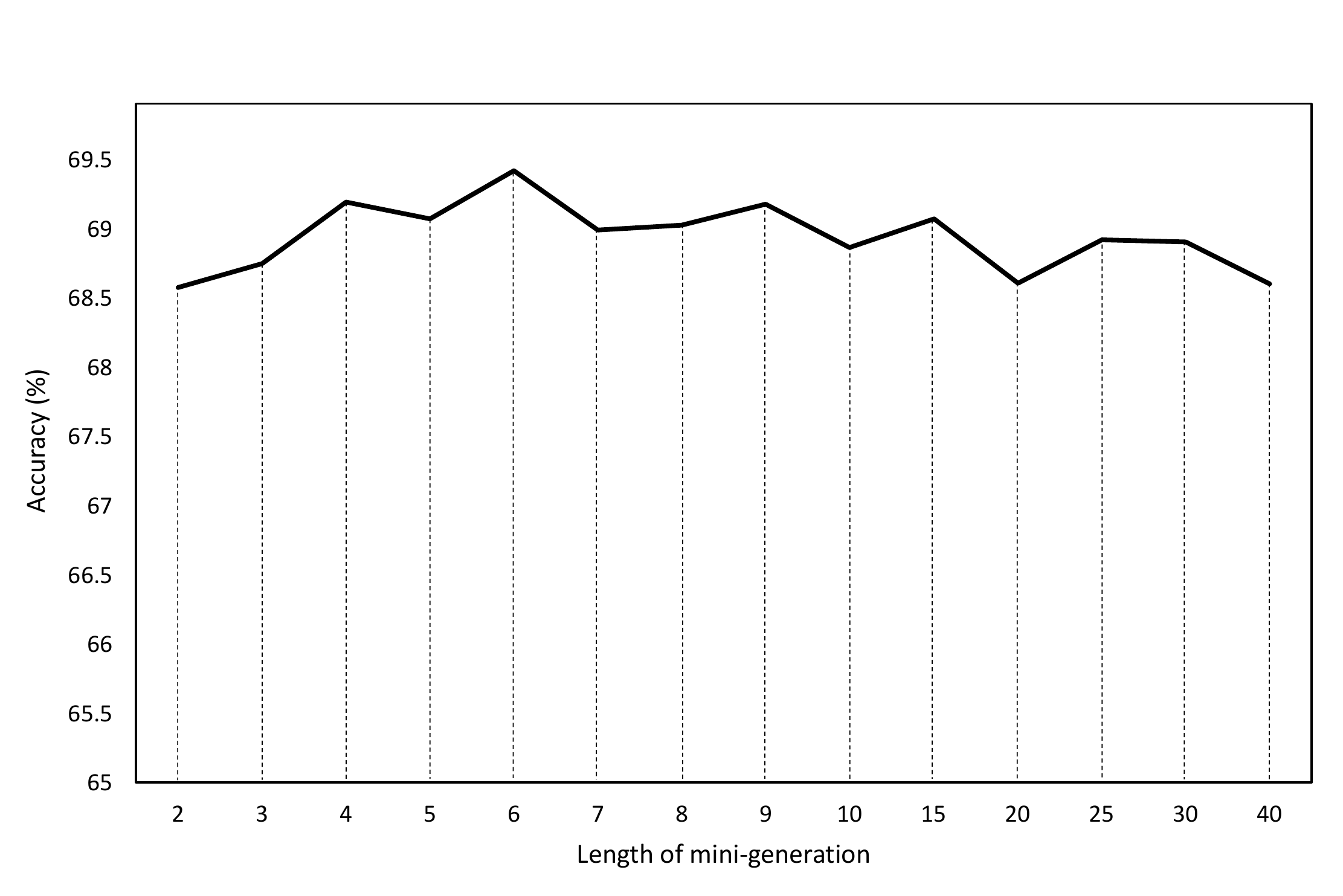}
\caption{Influence of different lengths of mini-generations.}
\label{mini_generation}
\end{figure}
\paragraph{Sensitivity Analysis.} Teacher--student differences are closely related to the length of each mini-generation. Thus, it is necessary to investigate what the best choice for the length of each mini-generation is. We conducted a sensitivity analysis on the length of each mini-generation on CIFAR100 using ResNet-20. As shown in Figure~\ref{mini_generation}, LSTSD improves as the length of the mini-generation increases from $1$ to $6$, and gradually declines as the length increases from $6$ to $40$. This is because a too short mini-generation length cannot guarantee teacher--student difference, while a too long one may introduce teachers with too low quality, which might mislead the training process.

\section{Conclusions}
In this paper, we propose a novel training policy called Long Short-Term Sample Distillation to train neural networks while relying on previous training updates for improved supervision. Our method decomposes the teacher signal for each sample from the previous training process into a long-term signal and a short-term one. The long-term teacher signal provides a stable teacher signal and guarantees teacher--student differences, while the short-term one ensures high-quality teaching.
Additionally, each sample has unique teachers, enabling the model to learn from numerous teachers over the course of training. The experimental results demonstrate the effectiveness of leveraging a long-term teacher and short-term teacher simultaneously, and learning from numerous teachers at the same time.

\small
\bibliographystyle{aaai}
\bibliography{AAAI-JiangL.6680}

\end{document}